\documentclass[conference]{IEEEtran}
\ifCLASSINFOpdf
\else
\fi

\usepackage[utf8]{inputenc}
\usepackage{cite}
\usepackage{amsmath,amssymb,amsfonts}
\usepackage{algorithmic}
\usepackage{graphicx}
\usepackage{textcomp}

\usepackage{color}
\usepackage{algorithm}
\usepackage{amsfonts}
\newcommand{\argmax}{\mathop{\rm arg~max}\limits}

\begin{document}
%
\title{Reinforcement Learning with Adaptive Curriculum Dynamics Randomization \\for Fault-Tolerant Robot Control}

\author{
\IEEEauthorblockN{Wataru Okamoto}
\IEEEauthorblockA{Graduate School of Science \\and Engineering, \\Chiba University\\
Chiba, Japan
}
\and
\IEEEauthorblockN{Hiroshi Kera}
\IEEEauthorblockA{Graduate School of Engineering, \\Chiba University \\
Chiba, Japan
}
\and
\IEEEauthorblockN{Kazuhiko Kawamoto}
\IEEEauthorblockA{Graduate School of Engineering, \\Chiba University \\
Chiba, Japan\\
e-mail: kawa@faculty.chiba-u.jp}}


%


\maketitle

\begin{abstract}
This study is aimed at addressing the problem of fault tolerance of quadruped robots to actuator failure, which is critical for robots operating in remote or extreme environments.
In particular, an adaptive curriculum
reinforcement learning algorithm with dynamics randomization (ACDR) is established.
The ACDR algorithm can adaptively train a quadruped robot 
in random actuator failure conditions and
formulate a single robust policy for fault-tolerant robot control.
It is noted that the \emph{hard2easy} curriculum is more effective than
the \emph{easy2hard} curriculum for quadruped robot locomotion.
The ACDR algorithm can be used to build a robot system that does not require additional modules for detecting actuator failures and
switching policies.
Experimental results show that the ACDR algorithm outperforms conventional algorithms in terms of the average reward and walking distance.
\end{abstract}


%
\IEEEpeerreviewmaketitle

\section{Introduction}
\label{sec:introduction}
Robots are being increasingly applied in a wide range of fields such as factory automation and disaster assessment and rescue.
In the existing approaches, the control law for robots are manually defined. 
However, in the presence of significant uncertainties in operating environments, the control law and states to be considered may not be easily identifiable, rendering such manual description challenging.
To address these issues, 
reinforcement learning has attracted attention as a promising method that can automatically establish the control law without a clear description of how to realize a task of interest.
A reinforcement learning framework learns, by trial and error, a policy that maximizes the \textit{reward}, which is a measure of the goodness of the behavior in a certain environment.
To realize reinforcement learning, it is necessary to only specify a task to be performed and design an appropriate reward function for the task.

Notably, to implement reinforcement learning in the real world, trial and error processes must be performed in real time.
However, this approach may pose safety risks for the robot and surrounding environment.
To ensure safety and cost effectiveness, robots are often trained in simulation environments before being applied in real world (Sim2Real). 
Reinforcement learning in simulation environments has achieved human-level performance in terms of the Atari benchmark~\cite{mnih2015human_Atari} and robot control~\cite{schulman2015trust_TRPO,levine2016end_roboticsReinforcementExample}.
However, Sim2Real is often ineffective owing to the gap between the simulation and real world, named \emph{reality gap}, which  
can be attributed to the differences in physical quantities such as friction and inaccurate physical modeling.
A straightforward approach to solve this problem is system identification, in which  
mathematical models or computer simulators of dynamical systems are established via measured data.
However, the development of a realistic simulator is extremely costly.
Even if system identification can be realized, many real-world phenomena associated with failure, wear, and fluid dynamics cannot be reproduced by such simulators.
A typical approach to bridge the reality gap is domain randomization~\cite{tobin2017domain}.
Domain randomization randomizes the parameters of the simulators to expose a model of interest
to various environments.
The model is expected to learn a policy that is effective in various environments, thereby narrowing the reality gap.
Domain randomization has been used to successfully apply Sim2Real for the 
visual servo control of robot arms~\cite{tobin2017domain, peng2018sim} and 
locomotion control of quadruped robots~\cite{tan2018sim}.

\begin{figure*}[!t]
  \begin{tabular}{c}
    \begin{minipage}[t]{175mm}
      \centering
      \includegraphics[width=\linewidth]{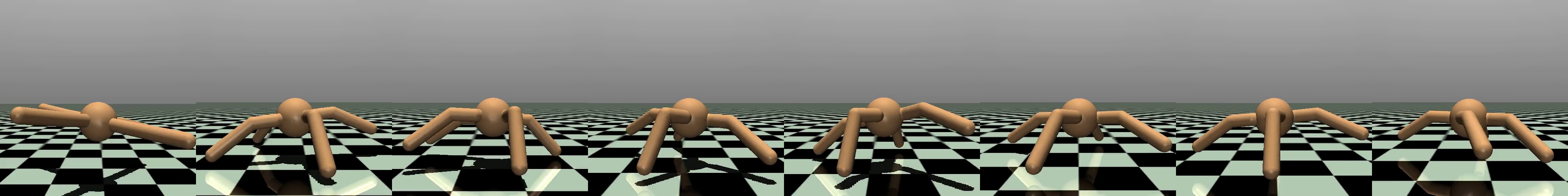}
    \end{minipage}\\
    \begin{minipage}[t]{175mm}
      \centering
      \includegraphics[width=\linewidth]{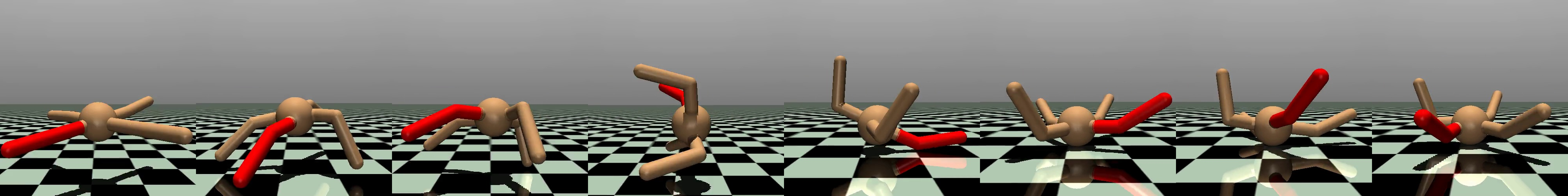}
    \end{minipage}
  \end{tabular}
  \caption{Illustration of quadruped robot locomotion with actuator failure.
  (Top) A quadruped robot trained by basic reinforcement learning can exhibit locomotion.  
  (Bottom) The quadruped robot turns over if the actuator of a leg (red) fails.}
  \label{fig:walk}
\end{figure*}

Notably, reinforcement learning involves certain limitations. Figure~\ref{fig:walk} shows that even though
a quadruped robot can walk owing to reinforcement learning (top row),
the robot may turn over owing to 
the actuator failure of one leg (bottom row). 
In this context, fault-tolerant control is critical for robots working in remote or extreme environments
such as disaster sites~\cite{robotInDisaster} because 
such robots can often not be repaired onsite.
Several researchers have focused on fault-tolerant robot control based on reinforcement learning
\cite{yang2020fault-aware,kume2017map_pfn}. 
However, these studies assume that the robots can detect internal failures and switch
their policies according to the failures.
Such robots require an additional module for diagnosing internal failures along with a module for control. 
In this context, robot systems designed for detecting all possible failures may be prohibitively complex~\cite{FaultsDiagnosisInRobotSystems_Review}.
To address this aspect, in this study, we establish a reinforcement learning algorithm for fault-tolerant control
of quadruped robots against actuator failure.
In this framework, the control law can ensure the locomotion of quadruped robots even if several actuators
in the robot legs fail.
Furthermore, we develop a fault-tolerant control system that does not include 
any self-diagnostic modules for actuator failures.
To this end, 
an adaptive curriculum dynamics randomization (ACDR) algorithm is proposed that can learn a single robust control
policy against actuator failures by
randomizing the dynamics of quadruped robots
with actuator failures.
In addition, we develop a curriculum learning algorithm that adaptively trains
the quadruped robots to achieve an enhanced walking ability.
It is noted that the \emph{hard2easy} curriculum, in which the training  is initiated in difficult conditions that progressively
becomes easier, is more effective than an \emph{easy2hard} curriculum, which is commonly used.
Experiments are conducted to demonstrate the effectiveness of the ACDR algorithm in comparison with
conventional algorithms.

The key contributions of this study can be summarized as follows:
\begin{itemize}
    \item We establish a reinforcement learning algorithm to address the problem of fault tolerance of quadruped robots to actuator failures
    and realize fault-tolerant control.
    \item The ACDR algorithm to proposed effectively learn a single robust control policy
    against actuator failures.
    The ACDR algorithm does not require any self-diagnosis modules for detecting internal failures,
    which must be implemented in conventional systems.
    \item We demonstrate that the ACDR algorithms outperform conventional algorithms in terms of the average reward
    and walking distance.
    \item For fault-tolerance control of quadruped robots, we find that
    the \emph{hard2easy} curriculum is effective than the \emph{easy2hard} curriculum.
\end{itemize}

\section{Related work}
\label{sec:RelatedWork}
Domain randomization~\cite{tobin2017domain, heess2017emergence} is aimed at enhancing generalization performance of the policy identified in simulations.
Domain randomization requires a prescribed set of $N$ dimensional simulation parameters, $\xi \in \Xi \subset \mathbb{R}^{N}$, to randomize and the corresponding sampling interval.
The $i$-th element of $\xi$ is sampled from the uniform distribution  $\mathrm{Uni}(L,U)$ on the closed interval $[L,U]$.
Using a set of the parameters $\{\xi\}$, we can realize robot learning in various simulation environments.
This randomization is expected to bridge the reality gap between the simulation and real world.
Tobin et al.~\cite{tobin2017domain} successfully applied Sim2Real for visual robot control by randomizing the visual rendering of the environments.
Moreover, Sim2Real has been successfully applied to control a robot arm~\cite{peng2018sim} and quadruped robot~\cite{tan2018sim} by randomizing physical parameters such as the robot mass and floor friction coefficient.
In particular, we refer to the physical parameter randomization ~\cite{peng2018sim,tan2018sim} as dynamics randomization, thereby distinguishing it from visual domain randomization~\cite{tobin2017domain}.
To implement dynamics randomization,
the interval $[L,U]$ must be carefully determined, as
an inappropriate interval is likely to result 
in failure to learn~\cite{Reinforcement_Okamoto}, 
a policy with considerable variance~\cite{ADR}, or a conservative policy~\cite{abdolhosseini2019learning_learningLocomotionSymmetry_master}.

For reinforcement learning, it is effective to start with an easy task in order to succeed in a difficult task.
Curriculum learning~\cite{bengio2009curriculum_curriculumLearning} is a learning process in which the difficulty of a task is gradually increased during training.
Curriculum learning is mainly applied in navigation~\cite{DeepWalk} and pickup tasks involving robot arms\cite{Kilinc2020FollowTO}.
Luo et al.~\cite{luo2020acceleratingReinforcementLearning_Curriculum} proposed a precision-based continuous curriculum learning~(PCCL) strategy that adjusts the required accuracy in a sequential manner during learning for a robot arm and can accelerate learning and increase the task accuracy.
In general, it is difficult to determine the appropriate level of difficulty of a task in curriculum learning.
Consequently, approaches that adaptively adjust the difficulty level for the agent through the learning process are attracting attention.
Wang et al.~\cite{POET} proposed a paired open-ended trailblazer~(POET) strategy, which ensures an appropriate level of difficulty by co-evolving the environment and policies, and enables walking in difficult terrain.
Moreover, curriculum learning in which the randomization interval is adaptively varied can help realize domain randomization~\cite{ADR, 2020-ALLSTEPS}.
Xie et al.~\cite{2020-ALLSTEPS} considered a stepping-stone walking task with constrained scaffolding.
The authors estimated the agent performance on each grid of the random parameters and adaptively varied the difficulty by decreasing and increasing the probability of occurrence of grids with high and low performance values, respectively.
This algorithm can enhance the performance and robustness of the model.
In general, the design of the environment and curriculum considerably influence the performance of learning to work~\cite{reda2020learning_SIGGRAPH_learningToLocomote}.
To realize natural walking behavior, it is necessary to limit the torque. To this end, a curriculum that starts with a higher value, such as $1.6$x, in the early stages of learning and gradually reduces the torque to the target value is considered to be effective~\cite{abdolhosseini2019learning_learningLocomotionSymmetry_master}.

To realize fault-tolerant robot control, 
Yang et al.~\cite{yang2020fault-aware} considered the joint damage problem in manipulation and walking tasks.
The authors established a method to model several possible fault states in advance and switch among them during training.
When the system encountered a failure, the method increased the robustness against joint damage by switching the policy.
Kume et al.~\cite{kume2017map_pfn} proposed a method for storing failure policies in a multidimensional discrete map in advance.
When the robot recognized a failure, it retrieved and applied these policies, thereby exhibiting a certain adaptability to robot failure.
Notably, the existing failure-aware learning methods assumed that the robot can recognize its damage state.
It remains challenging to develop
a system that can achieve high performance with a single policy regardless of whether the robot is in a normal or failed state, 
even if the robot does not recognize the exact damage state.

\begin{figure}[!t]
    \begin{algorithm}[H]
        \caption{Adaptive curriculum dynamics randomization}
        \label{algo:acdr}
        \begin{algorithmic}
        \REQUIRE $L,U$
        \REQUIRE $D$: Performance data buffer
        \REQUIRE $m,g_{th}$: Threshold
        \REQUIRE $\Delta_L,\Delta_U$: Update step size
        \STATE $\theta \leftarrow $ random weights 
        \STATE $D \leftarrow \phi$
		\WHILE{not~done}
		    \STATE $k \sim \mathrm{Uni}(L,U)  $ 
		    \STATE $l \sim \mathrm{Uni}\{0,1,2,3 \}$~choose leg
		    \STATE $o_{l,t} \leftarrow ko_{l,t-1} $ update $l$-th actuator by $k$
		    \STATE Run policy $\pi_{\theta}$ in environment with $k$
		    \STATE Optimize $\theta$ with PPO
		    \STATE Generate rollout $\tau = (s_0,a_0,\dots,s_T)$ 
		    \STATE $g \leftarrow 0$
		    \FOR{$s_t,a_t$ in $\tau$}
		        \STATE $g \leftarrow g + r(s_t,a_t)$ 
		    \ENDFOR
		    \STATE $D \leftarrow D \cup \{g\}$ \COMMENT{Add performance to buffer}
		    \IF{ $|D|$ $\geq m$ }
		        \STATE $\bar{g} \leftarrow$ Average of $D$
		        \IF{$\bar{g} \geq g_{th}$}
		            \IF{curriculum is \emph{easy2hard}}
		              \STATE $U \leftarrow U - \Delta_U$
		              \STATE $L \leftarrow L - \Delta_L$
		            \ELSIF{curriculum is \emph{hard2easy}}
		              \STATE $U \leftarrow U + \Delta_U$
		              \STATE $L \leftarrow L + \Delta_L$
		            \ENDIF
		            \STATE $g_{th} \leftarrow \bar{g}$
		        \ENDIF
		        \STATE $D \leftarrow \phi$ 
		    \ENDIF
		\ENDWHILE
        \end{algorithmic}
    \end{algorithm}
\end{figure}

\section{Method}
To realize fault-tolerant robot control without additional modules, we propose an ACDR algorithm for
reinforcement learning.
The objective is to formulate a single robust policy that can enable a robot to perform a walking task
even if actuator failure occurs.
We assume that the quadruped robot cannot recognize its damage state.
The ACDR algorithm can adaptively change the interval of the random dynamics parameter to efficiently train the robot.

\subsection{Reinforcement Learning}
We focus on the standard reinforcement learning problem~\cite{reinforcementLearning}, in which an agent interacts with its environment according to a policy to maximize a reward.
The state space and action space are expressed as $\mathbf{S}$ and $\mathbf{A}$, respectively.
For state $s_t \in \mathbf{S}$ and action $a_t \in \mathbf{A}$ at timestep $t$, the reward function $r:\mathbf{S} \times \mathbf{A} \rightarrow \mathbb{R}$ provides a real number that represents the desirability of performing a certain action in a certain state.
The goal of the agent is to maximize the multistep return $R_t = \sum_{t^\prime=t}^{T} \gamma^{t^\prime-t}
r(s_{t^\prime},a_{t^\prime})$ after $T$ steps, where $\gamma \in [0,1]$ is the discount coefficient, which indicates the importance of the future reward relative to the current reward.
The agent decides its action based on the policy $\pi_{\theta}:\mathbf{S}\times \mathbf{A}\rightarrow [0,1]$.
The policy is usually modeled using a parameterized function with respect to $\theta$.
Thus, the objective of learning is to identify the optimal $\theta^{*}$ as 
\begin{equation}
    \theta^{*} = \argmax_{\theta} J(\theta)
\end{equation}
where $J(\theta)$ is the expected return defined as
\begin{equation}
    J(\theta) = \mathbb{E}_{\tau \sim p_{\theta}(\tau)}\left[ \sum_{t=0}^{T-1} r(s_t, a_t) \right]
    \label{equ: expected return}
\end{equation}
where 
\begin{equation}
    p_{\theta}(\tau) = p(s_0)\prod_{t=0}^{T-1}p(s_{t+1} | s_t, a_t)\pi_{\theta}(s_t, a_t)
\end{equation} represents the joint probability distribution on 
the series of states and actions $\tau=(s_0, a_0, s_1, \dots, a_{T-1}, s_T)$ under policy $\pi_{\theta}$.

\subsection{Definition of Actuator Failure}
The actuator failure is represented through the variation in the torque characteristics of the joint actuators.
By denoting $o_{t} \in \mathbb{R}$ as the actuator output at timestep $t$,
we simply represent the output value of the failure state, $o_{t}^{\prime}$, as follows:
\begin{equation}
    o^{\prime}_t = ko_t.
    \label{eq:failure}
\end{equation}
where $k \in [0,1]$ is a failure coefficient.
Failure coefficients of $k=1.0$ and $k\not=1.0$ indicate that the actuator is in the normal or failed state, respectively.
A larger deviation of $k$ from $1.0$ corresponds to a larger degree of robot failure, and $k=0.0$ means that the actuator cannot be controlled.
Figure~\ref{fig:walk} illustrates an example of failure of an 8-DOF robot that had learned to walk.
The top row in Fig.~\ref{fig:walk} shows a normal robot with $k=1.0$, and the bottom row in Fig.~\ref{fig:walk} shows a failed robot (red leg damaged) with $k=0.0$.
Because the conventional reinforcement learning does not take into account environmental changes owing to robot failures, the robot falls in the middle of the task and cannot continue walking, as shown in the bottom row in
Fig.~\ref{fig:walk}.

\begin{figure}[!t]
    \centering
    \includegraphics[width=\linewidth]{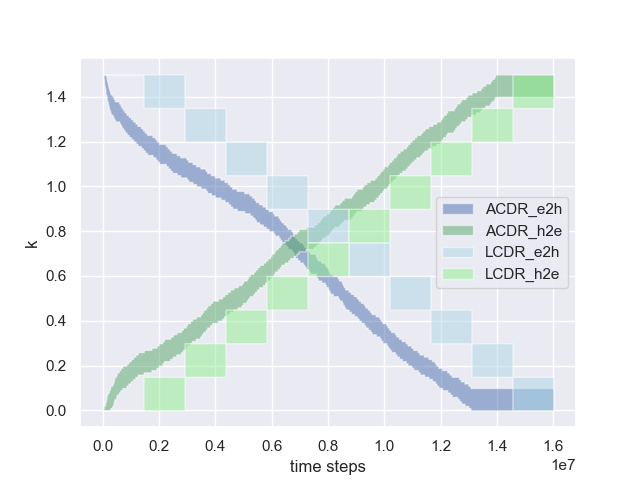}
    \caption{Time update of intervals $[L,U]$ of the failure coefficient $k$ in curriculum learning. The intervals are represented by the colored regions.
    ACDR\_e2h and ACDR\_h2e represent \emph{easy2hard} and \emph{hard2easy} curricula for ACDR, respectively. 
    LCDR\_e2h and LCDR\_h2h represent linear \emph{easy2hard} and \emph{hard2easy} curricula, respectively.}
    \label{fig:k_flactuation}
\end{figure}

\subsection{Adaptive Curriculum Dynamics Randomization}
To implement fault-tolerant robot control
the ACDR algorithm employs dynamics randomization by randomly sampling the failure coefficient $k$
and curriculum learning by adaptively changing the interval of $k$.
The process flow of the ACDR algorithm is presented as Algorithm~\ref{algo:acdr} and can be described as follows.

For dynamics randomization, one leg $l$ of the quadruped robot is randomly chosen as
\begin{equation}
    l \sim \mathrm{Uni}\{0,1,2,3 \}
\end{equation}
where $\mathrm{Uni}\{0,1,2,3 \}$ is the discrete uniform distribution on the possible leg
index set $\{0,1,2,3\}$ of the quadruped robot.
The leg $l$ is broken at the beginning of each training episode, and
the failure coefficient $k$ is randomly chosen as 
\begin{equation}
k \sim \mathrm{Uni}(L,U)
\label{equ: cud}
\end{equation}
where $\mathrm{Uni}(L,U)$ is the continuous uniform distribution on the interval $[L,U]$.
According to Eq.~(\ref{eq:failure}), the broken actuator of the leg $l$ outputs $o'_{l,t}=ko_{l,t}$ in each training
episode.
As the actuator output changes, the dynamics model $p(s_{t+1} | s_t, a_t)$ also changes during training.
This dynamics randomization means that a policy can be learned under varying dynamics models instead of being learned for a specific dynamics model. 
Under dynamics randomization,
the expected return 
is determined by taking the expectation over $k$, as indicated in Eq.~(\ref{equ: expected return}),
as
\begin{equation}
    \mathbb{E}_{k \sim \mathrm{Uni}(L,U)}\left[ \mathbb{E}_{\tau \sim p(\tau|k)} \left[ \sum_{t=0}^{T-1} r(s_t,a_t) \right] \right] \longrightarrow\max.
\end{equation}
By maximizing this expected return, 
the agent can learn a policy to adapt to the dynamics changes owing to robot failures.

For curriculum learning, the interval $[L,U]$ in Eq.~(\ref{equ: cud}) is adaptively updated according to the return earned by the robot in training.
As mentioned previously, a large interval $[L,U]$ is likely to result in a policy with large variance ~\cite{ADR} or a conservative policy~\cite{abdolhosseini2019learning_learningLocomotionSymmetry_master}.
Thus, a small interval $[L,U]$ is assigned to Algorithm~\ref{algo:acdr} as the initial value.
The interval $[L,U]$ is updated as follows:
First, a trajectory of states and actions $\tau = (s_0,a_0,\dots,s_T)$ is generated based on a policy
$\pi_\theta$, where $\theta$ is the policy parameter optimized using a proximal policy optimization (PPO) algorithm~\cite{PPO}.
Subsequently, the return $g$ of the trajectory $\tau$ is calculated by accumulating the rewards as
\begin{equation}
    g = \sum_{t=0}^{T-1}r(s_t,a_t).
\end{equation}
By repeating the parameter optimization and return calculation $m$ times, the average value of the returns $\bar{g}$ can be determined.
This value is used to evaluate the performance of the robot and determine whether the interval $[L,U]$ must be updated.
If the average $\bar{g}$ exceeds a threshold $g_{th}$, the interval $[L,U]$ is updated, i.e.,
the robot is trained with the updated failure coefficients.

In this study, we consider two methods to update the interval $[L,U]$. The manner in which the interval is updated during curriculum learning is illustrated in Fig.~\ref{fig:k_flactuation}.
In the first method, the upper and lower bounds of the interval $[L,U]$ gradually decrease as
\begin{equation}
\left\{\begin{array}{l}
    U \leftarrow U - \Delta_{U}\\
    V \leftarrow V - \Delta_{V}
    \end{array}
    \right.
\end{equation}
given the initial interval $[1.5,1.5]$.
In all experiments, we set $\Delta_{U}=\Delta_{V}=0.01$.
This decrease rule represents an \emph{easy2hard} curriculum because smaller values of $L$ and $U$
correspond to a higher walking difficult. For example, when $k=0$, the leg does not move.
The second method pertains to a \emph{hard2easy} curriculum, i.e.,
\begin{equation}
\left\{\begin{array}{l}
    U \leftarrow U + \Delta_{U}\\
    V \leftarrow V + \Delta_{V}
    \end{array}
    \right.
\end{equation}
We set the initial interval as $[0.0,0.0]$. In this state,
the robot cannot move the broken leg, and its ability to address this failure is gradually enhanced.
Notably, although the \emph{easy2hard} curriculum is commonly used in curriculum learning,
the \emph{hard2easy} curriculum is more effective than the
\emph{easy2hard} configuration in accomplishing the task.
The difference between the two frameworks is discussed in Section \ref{subsec: e2h h2e}.

\begin{figure}[!t]
    \centering
    \includegraphics[width=\linewidth]{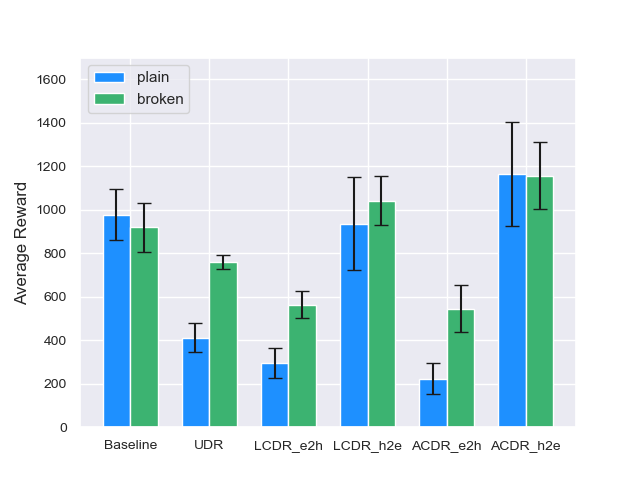}
    \caption{Average reward for the \emph{plain} ({blue}) and \emph{broken} ({green}) quadruped tasks. Error bars indicate the standard error. }
    \label{fig:AverageReward}
\end{figure}

\begin{figure}[!t]
    \centering
    \includegraphics[width=\linewidth]{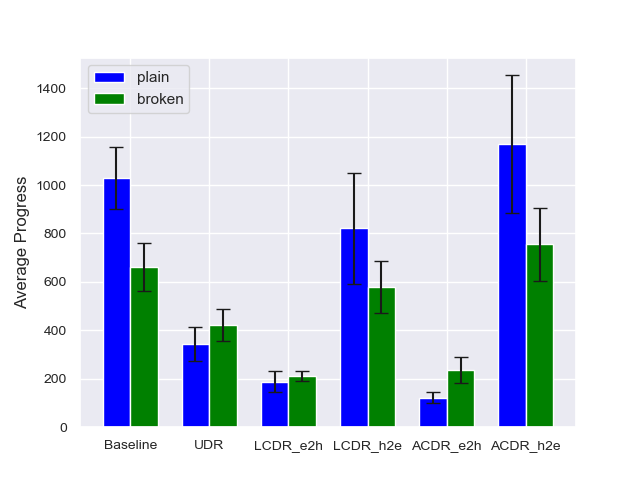}
    \caption{Average progress for the \emph{plain} ({blue}) and \emph{broken} ({green}) quadruped tasks. Error bars indicate the standard error. }
    \label{fig:AverageProgress}
\end{figure}

\subsection{Reward function}
In the analysis, we introduce a slightly modified reward function instead of using the commonly applied function described in \cite{rewardfunction},
because the existing reward function \cite{rewardfunction}
is likely to result in a conservative policy in which the robot does not walk but remains in place.
The existing reward function \cite{rewardfunction} can be defined by
\begin{equation}
    r = v_{\mathrm{fwd}} - 10^{-6}\| u \|^2 - 10^{-3}\|f_{\mathrm{impact}}\|^2 + 1,
    \label{equ: conventional reward func}
\end{equation}
where $v_{\mathrm{fwd}} = \frac{x^{\prime} - x}{\Delta t}$ is the walking speed, $u$ is the vector of joint torques, and $f_{\mathrm{impact}}$ is the contact force.
The reward function in Eq.~(\ref{equ: conventional reward func})
yields a reward of $1$ even when the robot falls or remains stationary without walking.
Therefore, this reward function is likely to lead to a conservative policy.
To ensure active walking of the robot, we modify the reward function as
\begin{equation}
r = v_{\mathrm{fwd}} - 10^{-6}\| u \|^2 - 10^{-3}\|f_{\mathrm{impact}}\|^2 + s,
\label{equ: our reward func}
\end{equation}
where 
\begin{equation}
s = 
    \begin{cases}
    0 & \text{if the robot is falling.}\\
    1 & \text{otherwise}
    \end{cases}
\end{equation}
Preliminary experiments demonstrate that the modified reward function in Eq.~(\ref{equ: our reward func})
can ensure more active walking of the robots.

\section{Experiments}
We conduct  experiments in the Ant-v2 environment provided by OpenAI Gym~\cite{openai}. 
This environment runs on the simulation software MuJoCo~\cite{mujoco}, and the task is aimed at realizing rapid advancement of the quadruped robot, as shown in Fig.~\ref{fig:walk}.
To realize reinforcement learning, we use the PPO algorithm~\cite{PPO}, which is suitable for high-dimensional continuous action environments.

\subsection{Experiment setup}
For reinforcement learning, we employ the actor and critic network 
consisting of two hidden layers of $64$ and $64$ neurons with 
$\tanh$ activation.
All networks use the Adam optimizer~\cite{Adam_kingmaadam}.
The related hyperparameters are listed in Table~\ref{tab: hyperparameters}.

\begin{table}[t]
    \centering
    \caption{Hyperparameters for experiments.}
    \begin{tabular}{c c}
    \hline
        Parameter & Value \\
        \hline \hline
        Learning rate & $0.00022$ \\
        Horizon & $128$ \\
        Minibatch size & $4$ \\
        epochs & $4$ \\
        Clipping parameter $\epsilon$ & $0.2$ \\
        Discount $\gamma$  & $0.99$ \\
        GAE parameter $\lambda$ & $0.95$ \\
        VF coefficient & $0.5$ \\
        Entropy coefficient & $0.01$ \\
    \hline
    \end{tabular}
    \label{tab: hyperparameters}
\end{table}

The ACDR algorithm is compared to the following three algorithms.
\textbf{Baseline}:
The baseline is a basic reinforcement learning algorithm for normal robots that implements neither dynamics randomization nor curriculum learning.
\textbf{Uniform Dynamics Randomization (UDR)}:
    UDR is a dynamics randomization algorithm that is based on the uniform
    distribution $\mathrm{Uni}(0.0,1.5)$ and does not implement curriculum learning.
\textbf{Linear Curriculum Dynamics Randomization (LCDR)}:
    In the LCDR algorithm,
    the interval $[L,U]$ is linearly updated. 
    Detail of the LCDR can be found in \ref{appendix: LCDR}.
    Similar to the ACDR, the LCDR uses
    the \emph{easy2hard} and \emph{hard2easy} curricula.
    The progress of the two curricula, the legends of which are "LCDR\_e2h" and
    "LCDR\_h2e", is shown in Fig.~\ref{fig:k_flactuation}.

We evaluate the algorithms in terms of the average reward and average walking distance in two conditions: \emph{plain} and \emph{broken}. 
In the \emph{plain}, the robots function normally, 
i.e., the failure coefficient is fixed at $k=1.0$.
In the \emph{broken}, the robots are damaged, and $k$ randomly
takes a value in $[0.0,0.5]$.
We implement each algorithm on the two conditions, with five random seeds for each framework. 
The average reward and average walking distance 
are calculated as the average of 100 trials for each seed.
The average walking distance is one that is suitable for evaluating
the walking ability of the robot.

\subsection{Comparative evaluation with conventional algorithms}
\label{sec:result}
The average reward and average walking distance for all algorithms
are shown in Fig.~\ref{fig:AverageReward} and 
Fig.~\ref{fig:AverageProgress}, respectively.
In both the \emph{plain} and \emph{broken} conditions, 
UDR is inferior to the Baseline, which does not consider robot
failures. This result indicates that the simple UDR cannot adapt to robot failures.
The \emph{hard2easy} curriculum of LCDR, denoted by LCDR\_h2e,
outperforms the Baseline in terms of the average reward, as shown in Fig.~\ref{fig:AverageReward}.
However,  LCDR\_h2e corresponds to an inferior walking ability compared to that of the Baseline for both \emph{plain} and \emph{broken} conditions, as shown in
Fig.~\ref{fig:AverageProgress}.
This result indicates that LCDR\_h2e implements a conservative policy that does not promote walking as active as that generated by the Baseline.
For both LCDR and ACDR, 
the \emph{hard2easy} curriculum outperforms the \emph{easy2hard} curriculum,
as discussed in Section \ref{subsec: e2h h2e}.
Moreover, ACDR\_h2e achieves the highest performance in all cases.
This finding indicates that ACDR\_h2e can avoid the formulation of a conservative policy
by providing tasks with difficulty levels that match the progress of the robot.

\begin{figure}[!t]
    \centering
    \includegraphics[width=\linewidth]{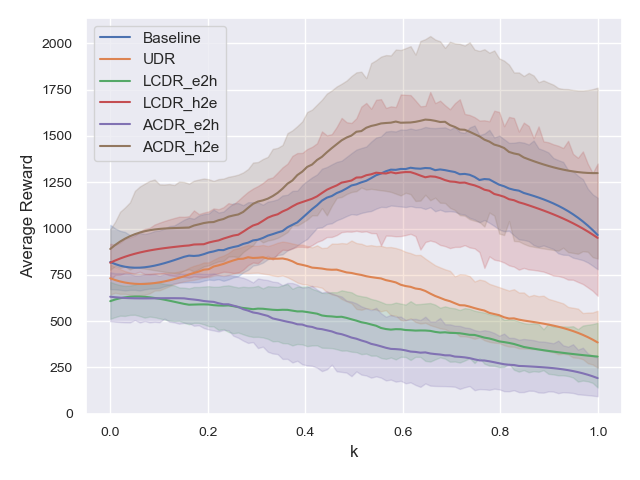}
    \caption{Average reward of all algorithms over various failure coefficients $k\in[0,1]$.}
    \label{fig:Generalization}
\end{figure}

Figure ~\ref{fig:Generalization} shows the average reward for 
each failure coefficient $k$ in the interval $[0,1]$. 
ACDR\_h2e earns the highest average
reward for any $k$, and ACDR\_h2e achieves the highest
performance in any failure state.
Moreover,
the performance of 
UDR, LCDR\_e2h, and ACDR\_e2h, deteriorate as $k$ increases.
This result indicates that the robots implementing these strategies cannot learn to walk.
Therefore, even when $k$ increases and it is easy to move the leg, 
the robots fall and do not earn rewards.

\subsection{Comparative evaluation with non-curriculum learning}\label{sec: non-cur}
\begin{figure}[!t]
    \centering
    \includegraphics[width=\linewidth]{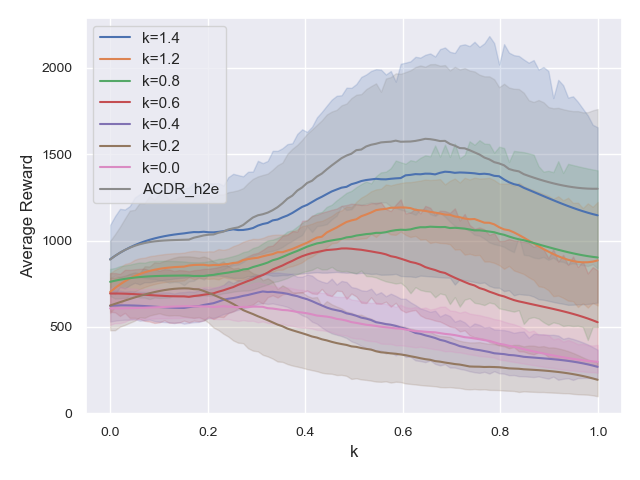}
    \caption{Average reward of the non-curriculum algorithm with fixed failure coefficient $k$ for various failure coefficients $k\in[0,1]$.}
    \label{fig:GeneralizationVariousK}
\end{figure}

The experiment results show that the \emph{hard2easy} curriculum is
effective against robot failures.
To evaluate whether the \emph{hard2easy} curriculum contributes to an enhanced performance, we compare this framework with a non-curriculum
algorithm in which the robot is trained with a fixed failure 
coefficient $k$. 
The values of $k$ are varied as 
$\{0.0, 0.2, 0.4, 0.6, 0.8, 1.2, 1.4\}$.
For each $k$,
Fig.~\ref{fig:GeneralizationVariousK} shows the average reward
earned by the robot on interval $[0,1]$ of the failure coefficient.
ACDR\_h2e still achieves the highest performance over the interval $[0,1]$.
Therefore, the \emph{hard2easy} curriculum can enhance the robustness against robot failures.


\subsection{Comparative evaluation between easy2hard and hard2easy curricula}
\label{subsec: e2h h2e}
The \emph{hard2easy} curriculum gradually decreases the degree of robot failure and makes it easier for the robot to walk.
In contrast, the \emph{easy2hard} curriculum gradually increases the degree of difficulty, and eventually, the leg stops moving.
Figures~\ref{fig:AverageReward} and \ref{fig:AverageProgress}
show that the \emph{hard2easy} curriculum is more effective than the \emph{easy2hard} curriculum. 
We discuss the reason below.

\begin{figure}[!t]
    \centering
    \includegraphics[width=\linewidth]{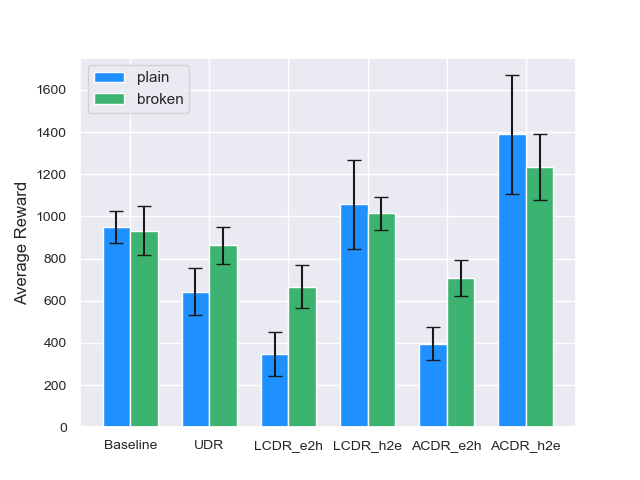}
    \caption{Average reward for the \emph{plain} ({blue}) and \emph{broken} ({green}) quadruped tasks. Error bars indicate standard error. The comparison algorithm is trained in the interval $k \in [0.5,1.5]$.}
    \label{fig:AverageReward_trainK0515}
\end{figure}

\begin{figure}[!t]
    \centering
    \includegraphics[width=\linewidth]{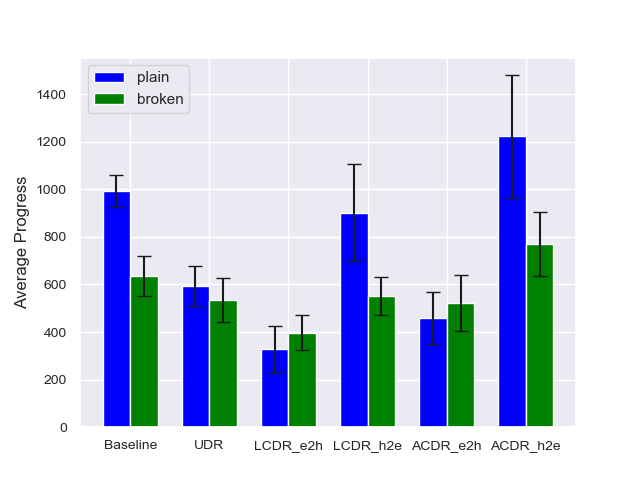}
    \caption{Average progress for the \emph{plain} ({blue}) and \emph{broken} ({green}) quadruped tasks. Error bars indicate the standard error. The comparison algorithm is trained in the interval $k \in [0.5,1.5]$.}
    \label{fig:AverageProgress_trainK0515}
\end{figure}

\begin{figure}[!t]
    \centering
    \includegraphics[width=\linewidth]{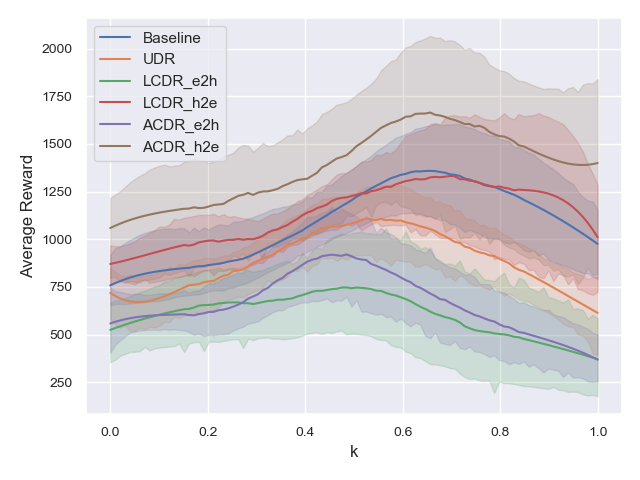}
    \caption{Average reward of all algorithms across various failure coefficients $k\in[0.5,1.5]$.}
    \label{fig:Generalization_trainK0515}
\end{figure}

Figure~\ref{fig:GeneralizationVariousK} shows that
the average reward tends to increase as $k$ increases
and deteriorates as $k$ decreases.
For example, $k=1.4$ corresponds to the highest average reward over the interval
$[0,1]$, whereas $k=0.0$ or $0.2$ yield low rewards.
This result suggests that it is preferable to avoid training at $k=0.0$ to allow the robot to walk even under failure.
Hence, we can hypothesize that the \emph{easy2hard} curriculum, in which training is performed at $k=0.0$ at the end of training, deteriorates the robot’s ability to walk.
To verify this hypothesis, we perform the same comparison as described in Section \ref{sec:result} 
for robots trained on the interval $k\in[0.5,1.5]$; in particular, in this framework, the robots are not trained at $k=0$.
Figures~\ref{fig:AverageReward_trainK0515} and \ref{fig:AverageProgress_trainK0515} show the average reward and average walking distance, respectively,
and Fig.~\ref{fig:Generalization_trainK0515} shows the average reward for each $k$
over the interval $[0,1]$.
Comparison of these results with those presented in
Figs.~\ref{fig:AverageReward}, \ref{fig:AverageProgress}, and \ref{fig:Generalization},
demonstrates that by avoiding training in the proximity of $k=0$, 
the average rewards of LCDR\_e2h and ACDR\_e2h are increased.
In summary, 
the \emph{easy2hard} curriculum deteriorates the learned policy owing to the strong degree of robot failure introduced at the end of training, whereas the \emph{hard2easy} curriculum does not.

Furthermore,
the average rewards of UDR, LCDR\_h2e, and ACDR\_h2e increase, 
as shown in Figs.~\ref{fig:AverageReward_trainK0515}, \ref{fig:AverageProgress_trainK0515},
and \ref{fig:Generalization_trainK0515}.
These results indicate that training in the proximity of $k=0$ is not necessary to achieve
fault-tolerant robot control.
Thus, interestingly, the robot can walk even when one of the legs cannot be effectively moved, without being trained under the condition.


\section{Conclusion}
This study was aimed at realizing fault-tolerant control of quadruped robots against actuator failures.
To solve the problem, we propose a reinforcement learning algorithm with 
adaptive curriculum dynamics randomization, abbreviated as ACDR.
We apply domain randomization to fault-tolerant robot control toward Sim2Real in robotics.
Furthermore, we developed an adaptive curriculum learning framework to enhance the effectiveness of domain randomization.
The ACDR algorithm could formulate a single robust policy to realize fault-tolerant robot control.
Notably, the ACDR algorithm can facilitate the development of a robot system that does not require
any self-diagnostic modules for detecting actuator failures.

Experiment results demonstrate that the combination of curriculum learning and dynamics randomization
is more effective than non-curriculum learning or non-randomization of dynamics.
In addition, the \emph{hard2easy} curriculum is noted to be
more effective than the \emph{easy2hard} curriculum.
This finding suggests that it is preferable to train quadruped robots in the order of difficulty opposite to that implemented in
standard curriculum learning.
In future work, there are several possible research directions of improving the ACDR algorithm.
A combination of automatic curriculum design (e.g. \cite{NEURIPS2020_emergent_PAIRED}) 
and domain randomization can be a promising direction.


\section*{Acknowledgment}
This work was supported by JSPS KAKENHI Grant Number JP19K12039.



%
\bibliographystyle{unsrt}
\bibliography{bib}

\begin{thebibliography}{10}

\bibitem{mnih2015human_Atari}
Volodymyr Mnih, Koray Kavukcuoglu, David Silver, Andrei~A Rusu, Joel Veness,
  Marc~G Bellemare, Alex Graves, Martin Riedmiller, Andreas~K Fidjeland, Georg
  Ostrovski, et~al.
\newblock Human-level control through deep reinforcement learning.
\newblock {\em nature}, 518(7540):529--533, 2015.

\bibitem{schulman2015trust_TRPO}
John Schulman, Sergey Levine, Pieter Abbeel, Michael Jordan, and Philipp
  Moritz.
\newblock Trust region policy optimization.
\newblock In {\em International conference on machine learning}, pages
  1889--1897. PMLR, 2015.

\bibitem{levine2016end_roboticsReinforcementExample}
Sergey Levine, Chelsea Finn, Trevor Darrell, and Pieter Abbeel.
\newblock End-to-end training of deep visuomotor policies.
\newblock {\em The Journal of Machine Learning Research}, 17(1):1334--1373,
  2016.

\bibitem{tobin2017domain}
Josh Tobin, Rachel Fong, Alex Ray, Jonas Schneider, Wojciech Zaremba, and
  Pieter Abbeel.
\newblock Domain randomization for transferring deep neural networks from
  simulation to the real world.
\newblock In {\em 2017 IEEE/RSJ International Conference on Intelligent Robots
  and Systems (IROS)}, pages 23--30. IEEE, 2017.

\bibitem{peng2018sim}
Xue~Bin Peng, Marcin Andrychowicz, Wojciech Zaremba, and Pieter Abbeel.
\newblock Sim-to-real transfer of robotic control with dynamics randomization.
\newblock In {\em 2018 IEEE international conference on robotics and automation
  (ICRA)}, pages 1--8. IEEE, 2018.

\bibitem{tan2018sim}
Jie Tan, Tingnan Zhang, Erwin Coumans, Atil Iscen, Yunfei Bai, Danijar Hafner,
  Steven Bohez, and Vincent Vanhoucke.
\newblock Sim-to-real: Learning agile locomotion for quadruped robots.
\newblock {\em arXiv preprint arXiv:1804.10332}, 2018.

\bibitem{robotInDisaster}
Ivana Kruijff-Korbayov{\'a}, Francis Colas, Mario Gianni, Fiora Pirri, Joachim
  de~Greeff, Koen Hindriks, Mark Neerincx, Petter {\"O}gren, Tom{\'a}{\v{s}}
  Svoboda, and Rainer Worst.
\newblock Tradr project: Long-term human-robot teaming for robot assisted
  disaster response.
\newblock {\em KI-K{\"u}nstliche Intelligenz}, 29(2):193--201, 2015.

\bibitem{yang2020fault-aware}
Fan Yang, Chao Yang, Di~Guo, Huaping Liu, and Fuchun Sun.
\newblock Fault-aware robust control via adversarial reinforcement learning.
\newblock {\em arXiv preprint arXiv:2011.08728}, 2020.

\bibitem{kume2017map_pfn}
Ayaka Kume, Eiichi Matsumoto, Kuniyuki Takahashi, Wilson Ko, and Jethro Tan.
\newblock Map-based multi-policy reinforcement learning: enhancing adaptability
  of robots by deep reinforcement learning.
\newblock {\em arXiv preprint arXiv:1710.06117}, 2017.

\bibitem{FaultsDiagnosisInRobotSystems_Review}
Muhamad Azhar, A~Alobaidy, Jassim Abdul-Jabbar, and Saad Saeed.
\newblock Faults diagnosis in robot systems: A review.
\newblock {\em Al-Rafidain Engineering Journal (AREJ)}, 25:164--175, 12 2020.

\bibitem{heess2017emergence}
Nicolas Heess, Dhruva TB, Srinivasan Sriram, Jay Lemmon, Josh Merel, Greg
  Wayne, Yuval Tassa, Tom Erez, Ziyu Wang, SM~Eslami, et~al.
\newblock Emergence of locomotion behaviours in rich environments.
\newblock {\em arXiv preprint arXiv:1707.02286}, 2017.

\bibitem{Reinforcement_Okamoto}
Wataru Okamoto and Kazuhiko Kawamoto.
\newblock Reinforcement learning with randomized physical parameters for
  fault-tolerant robots.
\newblock In {\em 2020 Joint 11th International Conference on Soft Computing
  and Intelligent Systems and 21st International Symposium on Advanced
  Intelligent Systems (SCIS-ISIS)}, pages 1--4, 2020.

\bibitem{ADR}
Bhairav Mehta, Manfred Diaz, Florian Golemo, Christopher~J Pal, and Liam Paull.
\newblock Active domain randomization.
\newblock In {\em Conference on Robot Learning}, pages 1162--1176. PMLR, 2020.

\bibitem{abdolhosseini2019learning_learningLocomotionSymmetry_master}
Farzad Abdolhosseini.
\newblock {\em Learning locomotion: symmetry and torque limit considerations}.
\newblock PhD thesis, University of British Columbia, 2019.

\bibitem{bengio2009curriculum_curriculumLearning}
Yoshua Bengio, J{\'e}r{\^o}me Louradour, Ronan Collobert, and Jason Weston.
\newblock Curriculum learning.
\newblock In {\em Proceedings of the 26th annual international conference on
  machine learning}, pages 41--48, 2009.

\bibitem{DeepWalk}
Diego Rodriguez and Sven Behnke.
\newblock Deepwalk: Omnidirectional bipedal gait by deep reinforcement
  learning.
\newblock In {\em 2021 IEEE International Conference on Robotics and Automation
  (ICRA)}, pages 3033--3039, 2021.

\bibitem{Kilinc2020FollowTO}
Ozsel Kilinc and G.~Montana.
\newblock Follow the object: Curriculum learning for manipulation tasks with
  imagined goals.
\newblock {\em ArXiv}, abs/2008.02066, 2020.

\bibitem{luo2020acceleratingReinforcementLearning_Curriculum}
Sha Luo, Hamidreza Kasaei, and Lambert Schomaker.
\newblock Accelerating reinforcement learning for reaching using continuous
  curriculum learning.
\newblock In {\em 2020 International Joint Conference on Neural Networks
  (IJCNN)}, pages 1--8. IEEE, 2020.

\bibitem{POET}
Rui Wang, Joel Lehman, Jeff Clune, and Kenneth~O. Stanley.
\newblock Paired open-ended trailblazer {(POET):} endlessly generating
  increasingly complex and diverse learning environments and their solutions.
\newblock {\em CoRR}, abs/1901.01753, 2019.

\bibitem{2020-ALLSTEPS}
Zhaoming Xie, Hung~Yu Ling, Nam~Hee Kim, and Michiel van~de Panne.
\newblock Allsteps: Curriculum-driven learning of stepping stone skills.
\newblock In {\em Proc. ACM SIGGRAPH / Eurographics Symposium on Computer
  Animation}, 2020.

\bibitem{reda2020learning_SIGGRAPH_learningToLocomote}
Daniele Reda, Tianxin Tao, and Michiel van~de Panne.
\newblock Learning to locomote: Understanding how environment design matters
  for deep reinforcement learning.
\newblock {\em CoRR}, abs/2010.04304, 2020.

\bibitem{reinforcementLearning}
Richard~S Sutton and Andrew~G Barto.
\newblock {\em Reinforcement learning: An introduction}.
\newblock MIT press, 2018.

\bibitem{PPO}
John Schulman, Filip Wolski, Prafulla Dhariwal, Alec Radford, and Oleg Klimov.
\newblock Proximal policy optimization algorithms.
\newblock {\em arXiv preprint arXiv:1707.06347}, 2017.

\bibitem{rewardfunction}
John Schulman, Philipp Moritz, Sergey Levine, Michael Jordan, and Pieter
  Abbeel.
\newblock High-dimensional continuous control using generalized advantage
  estimation.
\newblock {\em arXiv preprint arXiv:1506.02438}, 2015.

\bibitem{openai}
Greg Brockman, Vicki Cheung, Ludwig Pettersson, Jonas Schneider, John Schulman,
  Jie Tang, and Wojciech Zaremba.
\newblock Openai gym.
\newblock {\em arXiv preprint arXiv:1606.01540}, 2016.

\bibitem{mujoco}
Emanuel Todorov, Tom Erez, and Yuval Tassa.
\newblock Mujoco: A physics engine for model-based control.
\newblock In {\em 2012 IEEE/RSJ International Conference on Intelligent Robots
  and Systems}, pages 5026--5033. IEEE, 2012.

\bibitem{Adam_kingmaadam}
Diederik~P. Kingma and Jimmy Ba.
\newblock Adam: {A} method for stochastic optimization.
\newblock In Yoshua Bengio and Yann LeCun, editors, {\em 3rd International
  Conference on Learning Representations, {ICLR} 2015, San Diego, CA, USA, May
  7-9, 2015, Conference Track Proceedings}, 2015.

\bibitem{NEURIPS2020_emergent_PAIRED}
Michael Dennis, Natasha Jaques, Eugene Vinitsky, Alexandre Bayen, Stuart
  Russell, Andrew Critch, and Sergey Levine.
\newblock Emergent complexity and zero-shot transfer via unsupervised
  environment design.
\newblock In H.~Larochelle, M.~Ranzato, R.~Hadsell, M.~F. Balcan, and H.~Lin,
  editors, {\em Advances in Neural Information Processing Systems}, volume~33,
  pages 13049--13061. Curran Associates, Inc., 2020.

\end{thebibliography}

\appendices
\section{Linear Curriculum Dynamics Randomization}
\label{appendix: LCDR}
The LCDR updates the interval $[L,U]$ at the discrete
timestep $t=\Delta_{t},2\Delta_{t},\ldots, (N-1)\Delta_{t}$, where $\Delta_t$ is defined as 
\begin{equation}
 \Delta_{t} = \left\lfloor {T}/{N} \right\rfloor, 
\end{equation}
where $T$ is the predetermined total learning time, and $\lfloor\cdot\rfloor$ is the floor function.
We set $N=11$ in the experiments.
Similar to the ACDR, the LCDR implements two curricula: \emph{easy2hard} and \emph{hard2easy}.
Given the initial interval $[1.5,1.5]$,
the \emph{easy2hard} curriculum updates the interval as
\begin{align}
\begin{cases}
    L \leftarrow L - \Delta\, t\\
    U \leftarrow U - \Delta\, t
\end{cases}
\end{align}
where the update amount $\Delta$ is defined as
\begin{eqnarray}
    \Delta &=& \frac{k_{\max}}{N-1},
\end{eqnarray}
where $k_{\max}=1.5$ is the maximum value of the failure coefficient.
In addition, the \emph{hard2easy} curriculum updates the interval as 
\begin{align}
\begin{cases}
    L \leftarrow L + \Delta\, t\\
    U \leftarrow U + \Delta\, t
\end{cases}
\end{align}
given the initial interval $[0.0,0.0]$

\end{document}